\documentclass[10pt,twocolumn,letterpaper]{article}

\usepackage{cvpr}              % To produce the CAMERA-READY version
\usepackage{bbding}
\usepackage{float}
\usepackage{tabularx}
\usepackage{multirow}
\definecolor{cvprblue}{rgb}{0.21,0.49,0.74}
\usepackage[pagebackref,breaklinks,colorlinks,allcolors=cvprblue]{hyperref}
\usepackage{colortbl}
\definecolor{First}{rgb}{0.95, 0.62, 0.61}
\definecolor{Second}{rgb}{0.97,0.81,0.63}
\definecolor{Third}{rgb}{1.0, 0.97, 0.70}

\def\paperID{2505} % *** Enter the Paper ID here
\def\confName{CVPR}
\def\confYear{2025}

\title{MotionPRO: Exploring the Role of Pressure in Human MoCap and Beyond}

\author{
Shenghao Ren$^{*1}$, Yi Lu$^{*1}$, Jiayi Huang$^{1}$,  Jiayi Zhao$^{1}$, He Zhang$^{3}$, Tao Yu$^{3}$, Qiu Shen$^{\dag1,2}$, Xun Cao$^{1,2}$ \\
$^{1}$School of Electronic Science and Engineering, Nanjing University, Nanjing, China\\
$^{2}$Key Laboratory of Optoelectronic Devices and Systems with Extreme \\  Performances of MOE, Nanjing University, Nanjing, China \\
$^{3}$BNRist, Tsinghua University, Beijing, China\\
}

\begin{document}
% \maketitle

\twocolumn[{%
\renewcommand\twocolumn[1][!htb]{#1}%
\maketitle
\vspace{-9mm}
\begin{center}
    \centering
    \includegraphics[width=1\linewidth]{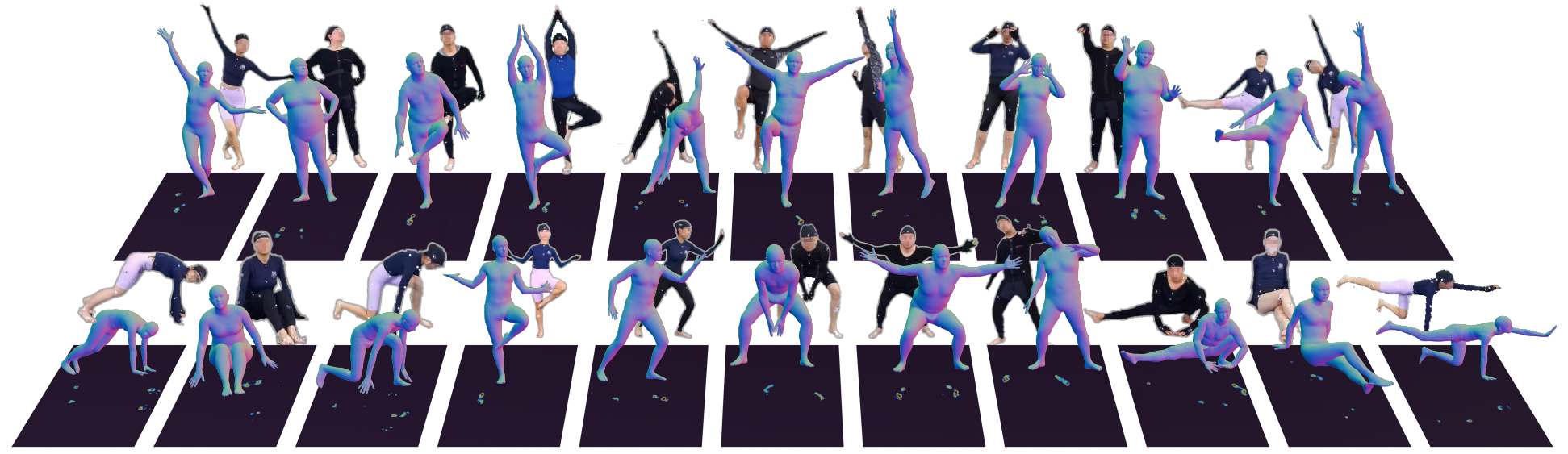}
    \vspace{-6mm}
    \captionof{figure}{\textbf{MotionPRO} is a large-scale human \textbf{Motion} capture dataset with \textbf{P}ressure, \textbf{R}GB  and \textbf{O}ptical sensors, which comprises 70 volunteers performing 400 types of motion, encompassing a total of 12.4M pose frames.}

    \label{fig:teaser}
    \vspace{-1mm}
\end{center}%
}]

\renewcommand{\thefootnote}{\fnsymbol{footnote}}
\footnotetext[1]{Equal contribution.}
\footnotetext[2]{Corresponding author.}
\renewcommand{\thefootnote}{\arabic{footnote}}

\begin{abstract}
\vspace{-2em}

Existing human Motion Capture (MoCap) methods mostly focus on the visual similarity while neglecting the physical plausibility.
As a result, downstream tasks such as driving virtual human in 3D scene or humanoid robots in real world suffer from issues such as timing drift and jitter, spatial problems like sliding and penetration, and poor global trajectory accuracy. 
In this paper, we revisit human MoCap from the perspective of interaction between human body and physical world by exploring the role of pressure.
Firstly, we construct a large-scale human \textbf{Motion} capture dataset with \textbf{P}ressure, \textbf{R}GB  and \textbf{O}ptical sensors (named \textbf{MotionPRO}), which comprises 70 volunteers performing 400 types of motion, encompassing a total of 12.4M pose frames. 
Secondly, we examine both the necessity and effectiveness of the pressure signal through two challenging tasks: (1) pose and trajectory estimation based solely on pressure: We propose a network that incorporates a small kernel decoder and a long-short-term attention module, and proof that pressure could provide accurate global trajectory and plausible lower body pose.  (2) pose and trajectory estimation by fusing pressure and RGB: We impose constraints on orthographic similarity along the camera axis and whole-body contact along the vertical axis to enhance the cross-attention strategy to fuse pressure and RGB feature maps. Experiments demonstrate that fusing pressure with RGB features not only significantly improves performance in terms of objective metrics, but also plausibly drives virtual humans (SMPL) in 3D scene. 
Furthermore, we demonstrate that incorporating physical perception enables humanoid robots to perform more precise and stable actions, which is highly beneficial for the development of embodied artificial intelligence. Project page is available at: \href{https://nju-cite-mocaphumanoid.github.io/MotionPRO/}{https://nju-cite-mocaphumanoid.github.io/MotionPRO/}
\end{abstract}    
\vspace{-1.5em}
\section{Introduction}

Human Motion Capture (MoCap) is a crucial foundation for motion understanding and imitation with diverse applications in AR/VR, humanoid robot actuation, and more. Current MoCap methods~\cite{kanazawa2018end, tripathi20233d, kolotouros2019learning, li2022cliff,fang2021reconstructing,li2021hybrik,goel2023humans,huang2018deep,yi2022physical,du2023avatars} have gained popularity due to their high precision in geometry similarity when evaluating the human body itself.

However, when applied to drive virtual humans in 3D scene or humanoid robots in real world, current human MoCap methods still exhibit dynamic inaccuracies, including temporal drift and jitter, as well as spatial issues such as sliding, floating, and penetration. This is because these methods mostly focus on the human individual, without considering the physical interaction with scene. This raises the question: Can we develop motion capture methods that incorporate dynamic interaction mechanisms? 

Our insight is that pressure signals can reflect the support provided by the ground to human body and even contain rich information on dynamic mechanisms and physics. The important role of pressure in pose estimation has been proved in the application of in-bed scene~\cite{clever20183d, clever2020bodies, wu2024seeing, yin2022multimodal, tandon2024bodymap, liu2022simultaneously}, but these methods cannot be generalized to daily motions. In this paper, we explore the role of pressure in human MoCap by constructing dataset, proposing baselines and conducting extended applications.

\begin{figure*}[htp]
\centering
\includegraphics[width=0.8\linewidth]{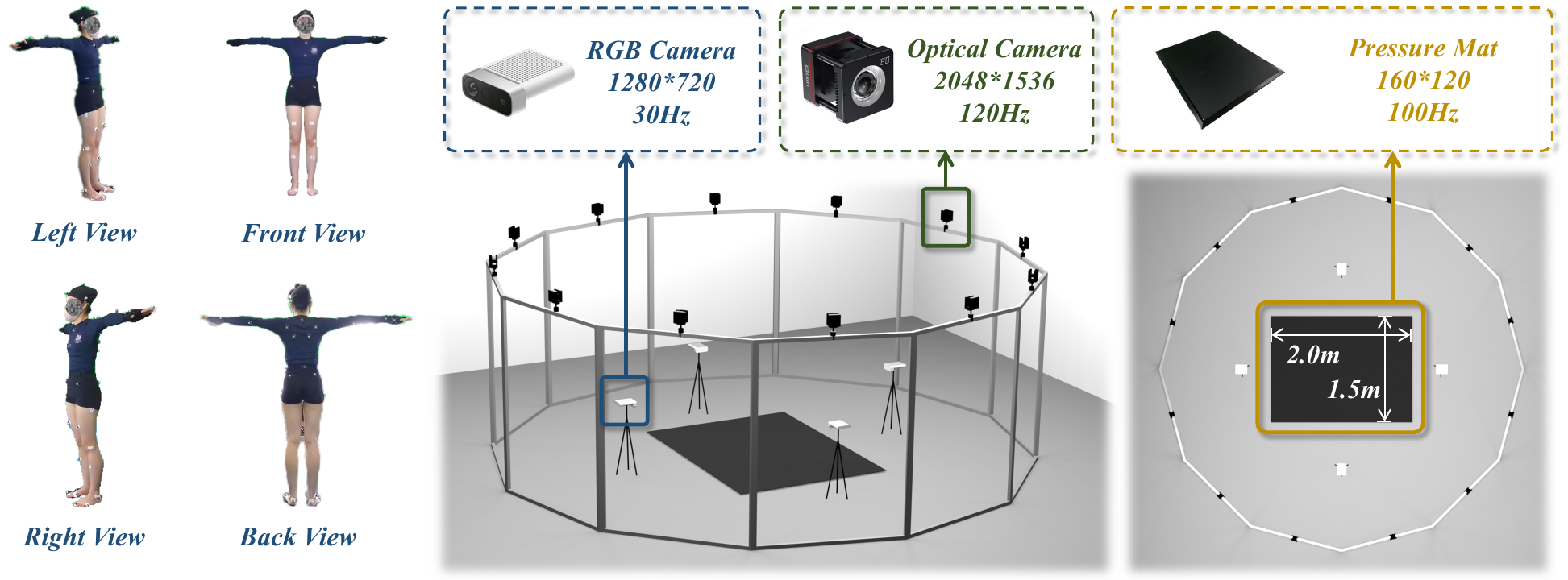}  
\caption{The architecture of our motion capture system for dataset collection. }  
\label{fig:system}   
\vspace{-1.5em}
\end{figure*}

Although there have been some pioneering datasets with pressure, such as MoYo~\cite{tripathi20233d} and PSU-TMM100~\cite{scott2020image}, they are limited to a single type of sport (i.e., Yoga and Taiji) and include fewer than 10 actors. We made significant efforts to construct a large-scale dataset, \textbf{MotionPRO}, which includes data from 70 volunteers (ages ranging from 17 to 61) performing 400 types of motion, encompassing a total of 12.4M pose frames. These motions span daily activities, traditional exercises, aerobic exercises, flexibility training, and specialized movements designed for humanoid robots. Specifically, we capture RGB videos from four perspectives, ground pressure signals of the whole human body, and position of 50 marker points of subjects from the optical MoCap system, as shown in Fig.~\ref{fig:system}. We follow~\cite{pavlakos2019expressive} to obtain highly accurate SMPL~\cite{loper2015smpl} annotations. 

We explore the potential necessity and effectiveness of pressure signals through two challenging tasks. Firstly,  we propose a pose and trajectory estimation method based solely on pressure by incorporating a small-kernel decoder and a long-short-term attention module. Experimental results proof that pressure could provide accurate global trajectory and plausible lower body pose, which is of great benefit to motion plausibility. Based on these findings, we further propose the \textbf{FRAPPE} baseline, which \textbf{F}uses \textbf{R}GB \textbf{A}nd \textbf{P}ressure for human \textbf{P}ose \textbf{E}stimation with plausible global translation. Aiming at combining the precise lower-body and global translation from pressure with the accurate local full-body pose from RGB, we impose constraints on orthographic similarity along the camera axis and whole-body contact along the vertical axis. 
Experiments demonstrate that FRAPPE outperforms SOTA RGB-based pose and trajectory estimation methods. 
% Notably, when the visual signal is occluded, our approach still performs well. 
Even when others suffer from extreme occlusions and vertical trajectory drift, ours remains physically plausible and accurate.

After evaluating our method in driving virtual humans (SMPL) in a 3D scene, we demonstrate that pressure can provide a physically plausible prior for human motion capture. This helps reduce jitter and drift over time while maintaining a realistic contact relationship with the ground. 
% These characters is the crucial for humanoid robot actuation. 
This happens to be the most critical issue in humanoid robot whole-body actuation. Thus, we conduct experiments on humanoid robots to further explore the role of pressure in robot actuation, specifically in improving the stability and precision of lower-body motion.

In this paper, we explore the crucial role of pressure in human motion capture. Our contributions are as follows:
\begin{itemize}
\item We construct \textbf{MotionPRO}, a large-scale human motion dataset with pressure, RGB, and optical sensors.

\item We propose \textbf{FRAPPE}, a baseline that fuses pressure and RGB data for precise and physically plausible pose and global trajectory prediction.

\item We conduct evaluations with different SOTA methods in both virtual human and humanoid robot to demonstrate the necessity and feasibility of our dataset and networks.

\end{itemize}

\section{Related Works}

\subsection{Vision-based Pose Estimation}

With the rapid development of deep learning, image-based human body pose estimation has developed rapidly~\cite{kanazawa2018end, tripathi20233d, kolotouros2019learning, li2022cliff, fang2021reconstructing, li2021hybrik, goel2023humans}.
However, RGB-based human pose estimation task is extremely ill-posed due to lack of depth information. 
In order to obtain accurate human body pose, people utilize the perspectives of temporal information~\cite{kocabas2020vibe, choi2021beyond, kanazawa2019learning, sun2019human, arnab2019exploiting}, prior knowledge of human body~\cite{bogo2016keep, kanazawa2018end, pavlakos2019expressive, zanfir2020weakly}, and precise camera model~\cite{li2022cliff, kissos2020beyond, kocabas2021spec, wang2023zolly}. However, these methods essentially place too much emphasis on alignment with 2D images, while neglecting global pose and trajectory in 3D scene.Recently, more and more works have focused on the task of estimating global trajectory. GLAMR~\cite{yuan2022glamr} utilizes local human poses and the relative relationships between humans, without considering the position of moving camera. SLAHMR~\cite{ye2023decoupling} and 
TRAM~\cite{wang2024tram} uses off-the-shelf SLAM algorithms to estimate camera trajectory. TRACE~\cite{sun2023trace} regresses human motion by utilizing optical flow between image frames. These methods ignore the most important relative relationship between human and the ground. While WHAM~\cite{shin2024wham} leverages contact label that is calculated only from foot velocity, which leads to low accuracy and a lack of consideration of the other joints' contact. 

\subsection{Pressure-based Pose Estimation}

 A single pressure frame is used for pose estimation of the lying pose~\cite{clever20183d, clever2020bodies}. In order to obtain higher quality pose estimation, ~\cite{yin2022multimodal, tandon2024bodymap, liu2022simultaneously, clever2022bodypressure} begin to explore extra information that pressure information cannot provide, such as depth, long wavelength infrared, and RGB. PIMesh~\cite{wu2024seeing} uses multi-frame pressure information as input and achieves higher-precision prone posture estimation by leveraging the distribution of pressure information over time.
 The auxiliary information derived from the direct pressure data can also serve as supplementary input for human pose estimation. The Center of Pressure (CoP)~\cite{tripathi20233d} is used to measure the stability of the estimated human pose. However, this constraint is only applicable to quasi-static movements. Foot contact~\cite{zhang2024mmvp} is also used to help constrain the relative position of people and the environment, but there is a absence of other body part contact. 
 
 Previous datasets with insole pressure sensors~\cite{han2023groundlink, mourot2022underpressure, scott2020image, zhang2024mmvp} capture foot pressure without whole-body pressure, while datasets containing whole-body pressure~\cite{liu2022simultaneously, clever2020bodies, wu2024seeing} are limited to lying pose. 
Additionally, Previous pressure datasets primarily focus on a limited range of motion types, such as Taiji~\cite{scott2020image}, Yoga~\cite{tripathi20233d}, lying in bed~\cite{liu2022simultaneously, clever2020bodies, wu2024seeing}, or a small subset of daily activities~\cite{han2023groundlink, mourot2022underpressure, zhang2024mmvp, luo2021intelligent}. The narrow scope of motion categories in these datasets restricts their generalizability, making them less suitable for diverse real-world applications. Thus, there is a need for a large-scale whole-body pressure dataset with diverse motion types.
\section{Dataset}

\begin{table*}[htp]
  \centering
  \resizebox{1\linewidth}{!}{
  \begin{tabular}{lccccccc}
    \toprule
    Datasets & Vision &Additional Sensor & Human Body & Subject & Pose Frames & Types  & Temporal\\
    \midrule

    Human3.6M~\cite{ionescu2013human3} & MV RGB  & - & SMPL & 11 & 3.6M & 15 & \checkmark\\ 
    
    % AGORA~\cite{patel2021agora} &  \\
    MPI-INF-3DH~\cite{mehta2017monocular} & MV RGB & - & Skeleton & 8 & 1.3M &$\sim$ 15 & \checkmark \\
    3DPW~\cite{von2018recovering} & SV RGB & IMU & SMPL & 7 & 51K & $\sim$ 8 & \checkmark \\
    
    \midrule

    GroundLink~\cite{han2023groundlink} & - & Force plate& SMPL & 7& 1.59M  & 19 &\checkmark\\

    UnderPressure~\cite{mourot2022underpressure} & - & Insole & Skeleton &10 & 2.02M & $\sim$8  & \checkmark\\
    PSU-TMM100~\cite{scott2020image} & DV RGB &Insole & Skeleton & 10& 1.36M & 24 & \checkmark\\

    MMVP~\cite{zhang2024mmvp} & SV RGBD &Insole& SMPL\&SMIL& 16 & 44K  & 10  & \checkmark\\

    SLP~\cite{liu2022simultaneously} & SV RGBD & Pressure mat (84*192), LWIR & 
 Keypoints & 109 & 14.7K & 15 & - \\
    PressurePose~\cite{clever2020bodies} & SV RGBD & Pressure mat (27*64) & SMPL & 20 & 207K & $\sim$6 &  - \\
    TIP~\cite{wu2024seeing} & SV RGBD & Pressure mat (40*56) & SMPL & 9 & 152K  & 30 &  \checkmark \\
    Intelligent Carpet~\cite{luo2021intelligent} & DV RGB & Pressure mat (96*96) & Skeleton & 10 & 180K & 15 & \checkmark\\
    MoYo~\cite{tripathi2023IP} & MV RGB &Pressure mat (37*110)& SMPL & 1 & 560K  & 82 & \checkmark\\
    \midrule
    Ours & MV RGB & Pressure mat (120*160) & SMPL & 70 & 12.4M & 400 & \checkmark\\

    \bottomrule
  \end{tabular}

  }
  \caption{Comparison of existing human motion capture datasets. S.V.: Single-V iew, M.V.: Multi-View, D.V.: Dual-View. SMPL\&SMIL means a mixture of SMPL and SMIL~\cite{hesse2018learning} (Skinned Multi-Infant Linear body model). IMU: Inertial Measurement Unit, LWIR: Long Wavelength Infrared camera. `-' indicates not included in the dataset. }
  \label{tab:multimodal}
  \vspace{-1.5em}
\end{table*}

To lay a solid foundation for exploring the role of pressure in MoCap, we have collected a large-scale dataset MotionPRO including Pressure, RGB, and Optical sensors.

\subsection{Setup and Configuration}
As shown in Fig.~\ref{fig:system}, the overall capture system comprises an Optical Motion Capture System, 4 RGB cameras, and a pressure mat. Specifically, the FZMotion Optical Motion Capture System~\cite{luster2024fzmotion}, equipped with 12 cameras, is employed to capture accurate human motion, with 50 reflective marker points placed on the human body. The pressure mat is positioned at the center of the MoCap cage, where motion capture quality is optimal. Surrounding the pressure mat, we position four consumer-grade cameras to capture front, side, and back body motion videos. The motion capture system, pressure mat, and RGB cameras collect data at frame rates of 120 Hz, 100 Hz, and 30 Hz, respectively. To unify the data frame rate across multiple sensors, we downsample all data to 30 Hz.
All motions must occur within the range of the pressure mat to ensure that each motion data point has a corresponding pressure measurement.

\subsection{Temporal and Spatial Alignment}

\textbf{Temporal synchronization.}
The time synchronization between the 12 optical cameras is completed by the FZMotion system via network cables. The four RGB cameras are configured in a daisy-chain setup, with the front RGB camera as the master device and the others as slave devices. Time synchronization between devices of the same type can be easily accomplished through the hardware's built-in synchronization method. As automatic time synchronization across different types of sensor hardware is not feasible, we manually synchronize the three data types using the volunteers stepping on the pressure mat as the beginning frame.

\textbf{Spatial alignment.}
In order to obtain the specific position of the camera, we place optical markers at the four top corners of the RGB camera. Using the optical mocap system, we can easily obtain the positions of the markers in the world coordinate system, and then compute the rotation $\boldsymbol{R}{w}^{c}\in SO(3)$ and translation $\boldsymbol{T}{w}^{c} \in \mathbb{R}^{3}$ of the RGB camera relative to the world coordinate system. Similarly, we can compute the rotation $\boldsymbol{R}{w}^{p}$ and translation $\boldsymbol{T}{w}^{p}$ of the pressure mat relative to the world coordinate system using the same method. Consequently, we can determine the relative position between any two cameras, as well as the relative position between any camera and the pressure mat. As shown in Fig.~\ref{fig:system}, we draw a 3D model of the overall system using precise relative positional relationships.

% \vspace{-0.5em}
\subsection{Motion Distribution}
% \vspace{-0.5em}

\begin{figure}[t]
\centering
\includegraphics[width=0.9\linewidth]{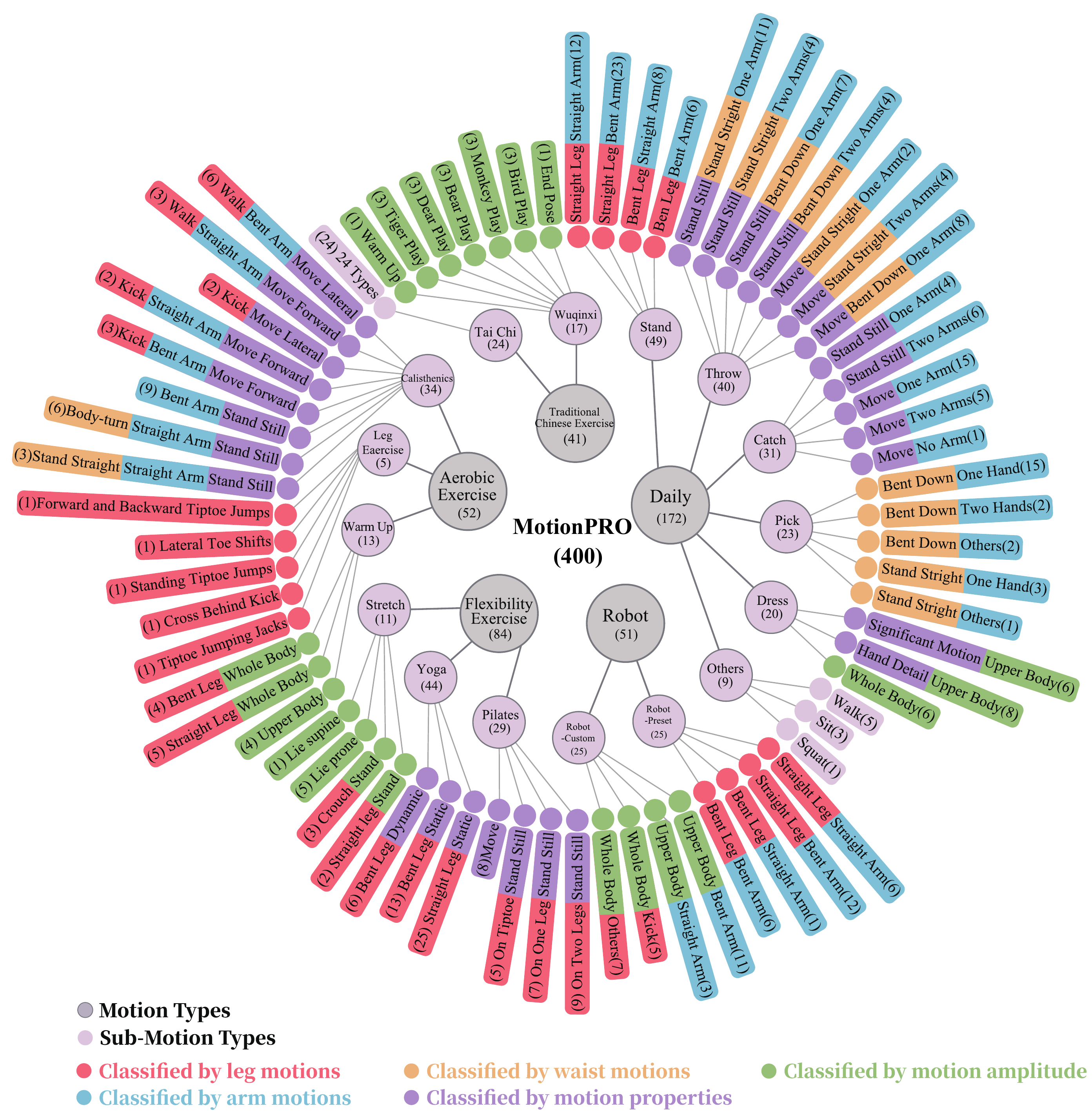}  
\caption{Hierarchal distribution of 400 motion types. }  
\label{fig:motion_type}
\vspace{-2em}
\end{figure}

As shown in Fig.~\ref{fig:motion_type}, MotionPRO encompasses a wide range of motion types, including daily motions, traditional exercise, aerobic exercise, flexibility exercise and special types designed for humanoid robot. It contains RGB camera videos and synchronized pressure data for a total of 729 sequences, amounting to 12.4M frames. We invite 70 volunteers of varying genders and diverse body types (with weights ranging from 44 kg to 109 kg and heights from 1.57 m to 1.84 m), aged between 17 and 61, ensuring variation and generalizability. All participants have consented to the use of their data for academic purposes. Tab.~\ref{tab:multimodal} provides statistics compared with other human MoCap datasets.

\subsection{Annotation Acquisition}
% Move to section 4
We use the SMPL~\cite{loper2015smpl} model as a representation of the human body. 
The SMPL model utilizes shape parameters \(\boldsymbol{\beta} \in \mathbb{R}^{10}\), pose parameters \(\boldsymbol{\theta} \in \mathbb{R}^{72}\) and a global translation \(\boldsymbol{T} \in \mathbb{R}^{3}\) as inputs. This model generates a triangulated mesh comprising 6,890 vertices.
The $k$ global joints of SMPL can be represented as $\boldsymbol{J}(\boldsymbol{\beta},\boldsymbol{\theta}, \boldsymbol{T}) \in \mathbb{R}^{k \times 3}$. 
Ground-truth SMPL~\cite{loper2015smpl} parameters are calculated from the mocap raw marker data by using Mosh++~\cite{mahmood2019amass}. 
To determine whether joint \( j \in \boldsymbol{J} \) is in contact with the pressure mat, we vertically project it onto the ground and calculate the sum of pressure in the vicinity as $  P_{j}$, as well as the distance to the ground plane $D_j$. We annotate the contact label $C_j$ by using the following strategy:

\begin{equation}
C_{j} = 
\left\{  
\begin{alignedat}{2}
1 \quad &\text{if} P_{j} \ge \tau_1 \ \text{and} \ D_j \le \tau_2\\
0 \quad &\text{otherwise,}
\end{alignedat}  
\right. 
\end{equation}
where \(\tau_1\) and \(\tau_2\) are thresholds for each variables.

\section{Baseline}

To evaluate the usefulness and significance of whole-body pressure in pose and global trajectory estimation, we investigate two challenging tasks: 1) pose and trajectory estimation using only pressure, and 2) pose and trajectory estimation by fusing pressure and RGB.

\subsection{Pose and Trajectory Estimation using Only Pressure}

The pressure signal is the combined force exerted by the human body on the ground under the action of gravity and vertical acceleration. In daily life, the pressure distribution between the body and the ground is highly sparse, meaning that the same pressure pattern from a single pressure image may correspond to thousands of possible human body motions. 
To address this ill-posed problem, we design a whole-body pose and trajectory estimator that relies solely on consecutive multiple pressure images.
Unlike~\cite{clever20183d, clever2020bodies}, which estimate pose directly from a single pressure image, we follow~\cite{luo2021intelligent, wu2024seeing} and use continuous multi-frame pressure images for pose estimation. Our approach aims to reduce the ambiguity in pose estimation by incorporating pressure information from adjacent frames.

\begin{figure}[t]
\centering
\includegraphics[width=1\linewidth]{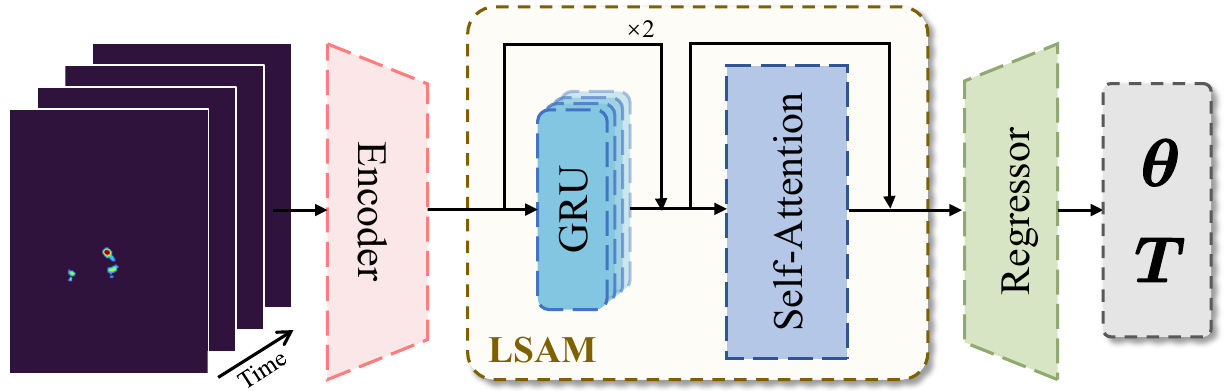}  
\caption{Pose and Trajectory estimation using only pressure.}  
\label{fig:pressure2smpl}   
\vspace{-1.5em}
\end{figure}

\begin{figure*}[htp]
\centering
\includegraphics[width=0.8\linewidth]{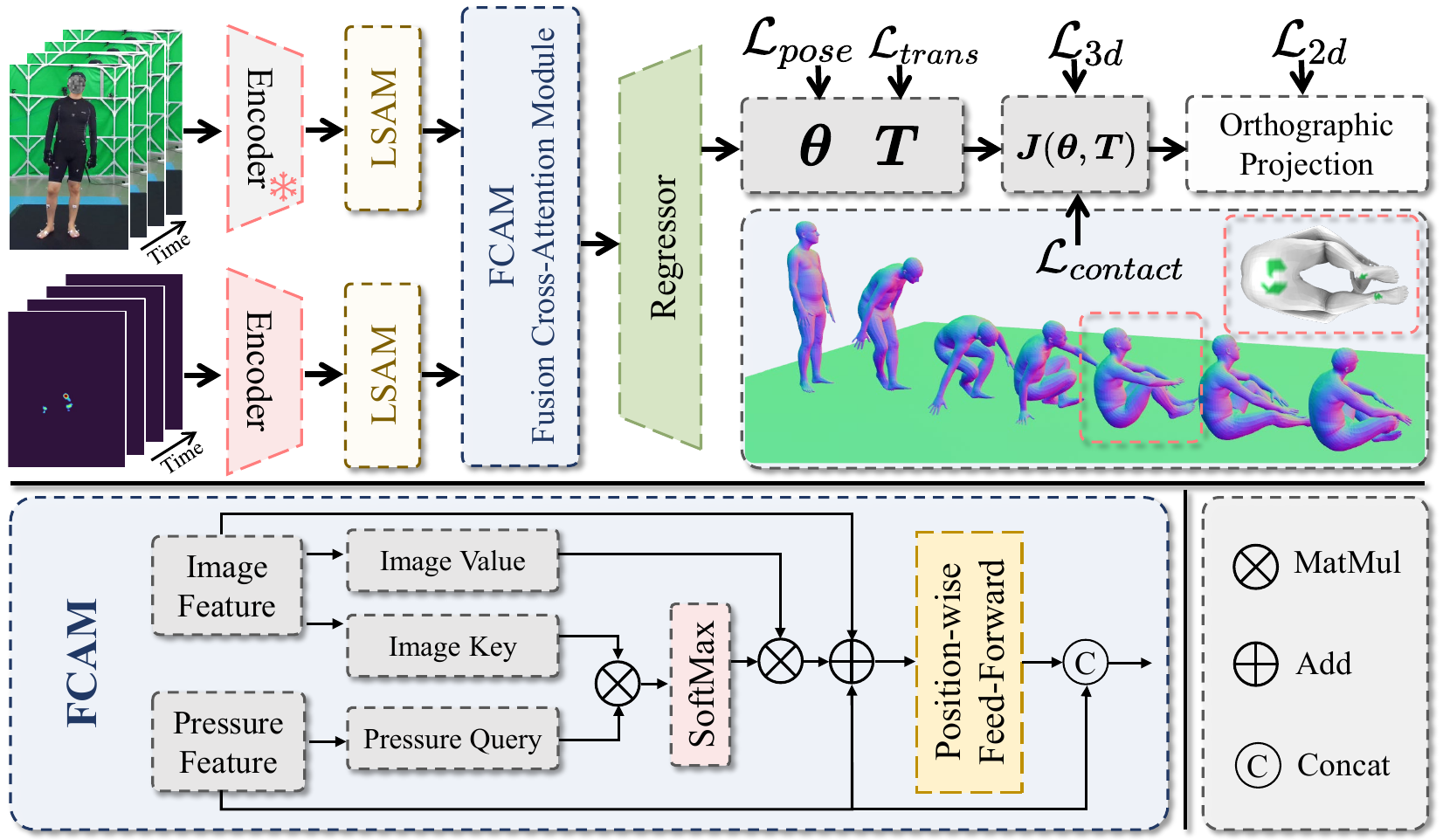}  
\caption{The framework of FRAPPE which fuses pressure and RGB for global pose and trajectory estimation.}  
\label{fig:method}   
\vspace{-1.5em}
\end{figure*}

As shown in Fig.\ref{fig:pressure2smpl}, our network consists of three parts: pressure encoder, temporal information processor, and human pose regressor. As useful information on the pressure image is extremely sparse when only feet are in contact with the ground, we reduce the convolution kernel size of ResNet~\cite{he2016deep} in an attempt to extract more refined pressure features. To fully exploit pressure features, we deisgn a Long and Short term Attention Module where GRU~\cite{chung2014empirical} extracts short-term contextual action, and self-attention~\cite{vaswani2017attention} extracts and enriches long-term dependencies through multiple attention heads. Finally, we follow~\cite{kocabas2020vibe} to construct the human pose regressor for the estimation of pose and translation parameters. Loss functions are as follows:

\vspace{-1.5em}
\begin{align}
 \mathcal{L}=
&\lambda_{pose}\mathcal{L}_{pose}+\lambda_{3d}\mathcal{L}_{3d}+ \\ \notag
&\lambda_{trans}\mathcal{L}_{trans} +\lambda_{contact}\mathcal{L}_{contact}, 
  \label{eq:pressure2smpl_loss}
\end{align}
% \vspace{-0.5em}
where $\lambda_{pose}$,$\lambda_{3d}$, $\lambda_{trans}$ , and $\lambda_{contact}$ are corresponding weights. The loss of pose parameters $\mathcal{L}_{pose}$ is the mean squared error between the predicted and ground-truth pose parameters. The 3D joint loss, \(\mathcal{L}_{3d}\), is the mean squared error between the predicted and ground-truth joint positions, after pelvis alignment. Global translation loss $\mathcal{L}_{trans}$ is the mean squared error between predicted and ground truth translation. The ground contact loss, \(\mathcal{L}_{contact}\), is the mean squared error between the predicted and the ground-truth global joints that are in contact with the ground.

\subsection{Pose and Trajectory Estimation by Fusing Pressure and RGB}

Pressure encodes the physical interaction between human body and ground, while RGB captures the horizontal vision signal of the human body. They play a complementary role in human motion estimation. Accordingly, we propose FRAPPE, a baseline that fuses pressure signal and monocular RGB images to obtain physically plausible motion. 

As shown in Fig.~\ref{fig:method} , FRAPPE add a RGB branch and a Fusion Cross-Attention Module (FCAM) compared to the pose estimate network only from pressure. The pressure branch and regressor remain the same. In the RGB image branch, we follow the previous work~\cite{kanazawa2019learning, kocabas2020vibe} and use a parameter-frozen pre-trained HRnet~\cite{sun2019deep} as the image encoder, which is proved by~\cite{li2022cliff}. To fully fuse features from two different domains, we utilize the cross-attention strategy to fuse precise visual geometry and physical dynamics. We set pressure feature as Query and image feature as Key and Value based on the belief that pressure contains more information related to the real physical world, such as contact and physical interaction.

The loss functions of FRAPPE are as follows:

\vspace{-1.5em}
\begin{align}
 \mathcal{L}_{FRAPPE}=
 % (\boldsymbol{\theta}, \boldsymbol{R},\boldsymbol{T})} = 
&\lambda_{pose}\mathcal{L}_{pose}+\lambda_{3d}\mathcal{L}_{3d} +\lambda_{2d}\mathcal{L}_{2d}  \\ \notag
&\lambda_{trans}\mathcal{L}_{trans}+\lambda_{contact}\mathcal{L}_{contact}, 
  \label{eq:frappe_loss}
\end{align}
% \vspace{-1.5em}
where $\lambda_{pose}$, $\lambda_{3d}$, $\lambda_{2d}$, $\lambda_{trans}$,  and $\lambda_{contact}$ are corresponding weights. The loss function of FRAPPE is consistent with that of pose estimation only from pressure except for $\mathcal{L}_{2d}$, which is the mean squared error of orthographic projection in the camera direction between the predicted joints and ground truth joints. 

Unlike most methods that use weak perspective projection camera model~\cite{kanazawa2018end, li2022cliff, goel2023humans}, we use orthographic projection camera model. 
As mentioned in~\cite{dwivedi2024tokenhmr} , there is a paradoxical decline in 3D pose accuracy with increasing 2D image alignment accuracy. We also observe that this contradiction exists not only in pose, but also in global trajectory due to the deception by the 2D image. In the weak perspective projection framework, the trajectory of the human body relative to the camera coordinate system is primarily aligned at the pixel level with the estimated human body in the 2D image, which leads to the coupling of pose and trajectory. At the same time, due to the depth ambiguity inherent in 2D images, the shape of the human body may change to accommodate the estimated translation in order to maintain 2D alignment. Obviously, weak perspective projection is useful when only local pose and shape are considered, and global trajectory is not taken into account. Hence, when focusing on global trajectory, orthogonal projection, which preserves scale in depth direction, becomes more effective.

\vspace{-0.5em}
\section{Experiments}
\noindent\textbf{Evaluation Metrics.} To evaluate our methods, we use the following metrics: MPJPE (Mean Per Joint Position Error), PMPJPE (Procrustes-aligned Mean Per Joint Position Error), PVE (Per Vertex Error), and Accel (Acceleration). 
For evaluating global trajectory, we utilize the GTraj (Global Trajectory error of root) and GMPJPE (Global Mean Per Joint Position Error) in pose and trajectory estimation only from pressure. For comparison with RGB-based methods, we follow~\cite{shin2024wham} and split sequences into segments of 100 frames and align each segment with ground truth by using the first two frames (WMPJPE) or all frames (WAMPJPE). Root Translation Error (RTE) over the entire trajectory, Jitter of moition, and Whole Body Contact Error (WBCE) are introduced for evaluation. WBCE represents the average absolute height distance between the ground plane and the joints which are in contact with ground. FS represents Foot Sliding during the contact. The unit of Jitter is $10m/s^2$. The unit of Accel is $m/s^2$. All other metrics are in $mm$.
 % \vspace{-1em}

\subsection{Pose and Trajectory Estimation using Only Pressure}
\vspace{-0.5em}

\begin{table}[tbp]
  \centering
  \resizebox{1\linewidth}{!}{
  \begin{tabular}{@{}lcccccc@{}}
    \toprule
    Methods & M. $\downarrow$ & LM. $\downarrow$ & PM. $\downarrow$ & LPM.  $\downarrow$ & GTraj $\downarrow$ & GM. $\downarrow$  \\
    \midrule
    IC~\cite{luo2021intelligent} & 299.3 & 239.7 & 205.1 & 94.9  & 403.2 & 357.7  \\
    IC[FT] & 133.0 & 94.9 &  100.8 & 45.5 & \cellcolor{First}{80.5} & 143.3 \\

    \midrule
    Ours  & \cellcolor{First}{90.6} & \cellcolor{First}{58.5} & \cellcolor{First}{70.2}& \cellcolor{First}{32.4}  & 85.9 & \cellcolor{First}{127.6} \\
    
    \bottomrule
  \end{tabular}
  }
  \caption{Evaluation of global pose and trajectory estimation only from pressure on MotionPRO. M.: MPJPE, PM. : PMPJPE, LM., LPM.: Lower body MPJPE, PMPJPE, GM.: GMPJPE. }
  \label{tab:pressure2pose_evaluation}
    \vspace{-2em}
\end{table}

We conduct an evaluation of our method against Intelligent Carpet (IC)~\cite{luo2021intelligent} and IC[FT] on MotionPRO. IC[FT] refers to training the Intelligent Carpet (IC) method on MotionPRO. As shown in Tab.~\ref{tab:pressure2pose_evaluation}, directly applying IC to our dataset leads to significant errors in both pose and global trajectory. This is because the IC dataset contains a limited range of motion types, noisy pressure data, and inaccurate annotation, which result in poor performance. Compared to IC[FT], we achieve significant improvements in pose estimation, though with a slight loss in global trajectory accuracy. Our method still retains an advantage in global motion estimation which is more important in real 3D scenes.

Benefiting from the contact loss, our method has a better performance on lower body pose estimation. This demonstrates that by extracting contact information and higher-dimensional physical interaction from pressure, the accuracy of lower body pose and plausible ground interaction along with accurate global trajectory can be ensured.

\subsection{Pose and Trajectory Estimation by Fusing Pressure and RGB}

\begin{table}[tbp]
  \centering
  \resizebox{1\linewidth}{!}{
  \begin{tabular}{@{}lccccc@{}}
    \toprule
    Methods & MPJPE $\downarrow$ & PMPJPE $\downarrow$ & PVE $\downarrow$  & Accel $\downarrow$  \\
    \midrule
    VIBE~\cite{kocabas2020vibe} & 59.7 & 40.9 & 82.9    & 19.6  \\
    CLIFF~\cite{li2022cliff} & 54.7 & 39.7 &  \cellcolor{Second}68.6  & 24.3 \\
    SMPLer-X~\cite{cai2024smpler} & \cellcolor{Second}{51.6} & {32.8}  & 72.4  & 437.3  \\
    TRACE~\cite{sun2023trace} & 61.4 & 43.2 &  81.4  & 14.6 \\

    WHAM~\cite{shin2024wham} & 160.4 & \cellcolor{First}{28.3}  & 227.5  & \cellcolor{First}2.9  \\
    PhysPT~\cite{zhang2024physpt} & 56.4 & 38.7  & 72.6  & \cellcolor{Second}{3.0}  \\
    \midrule
    Ours  & \cellcolor{First}{41.8} & \cellcolor{Second}30.2 & \cellcolor{First}58.6    & \cellcolor{Second}{3.0}  \\
    
    \bottomrule
  \end{tabular}
  }
  \caption{Evaluation of global pose estimation on MotionPRO.}
  \label{tab:frappe_pose}
\vspace{-1.5em}
\end{table}

\noindent\textbf{Evaluation in global pose estimation.} We compare FRAPPE with VIBE~\cite{kocabas2020vibe}, CLIFF~\cite{li2022cliff}, SMPLer-X~\cite{cai2024smpler}, TRACE~\cite{sun2023trace}, WHAM~\cite{shin2024wham} and PhysPT~\cite{zhang2024physpt} on the MotionPRO dataset. To evaluate global pose estimation, we align each orientation of compared methods with ground truth by using the first frame. 
As shown in Tab.~\ref{tab:frappe_pose}, FRAPPE outperforms almost all other methods in human pose estimation when considering global orientation. The reason why WHAM achieves a good PMPJPE but not a good MPJPE is that the global orientation estimated by WHAM varies over time. When driving virtual human or humanoid robot in 3D scene, we need a precise and reasonable global pose.

For qualitative comparison, as shown in Fig.~\ref{tab:frappe_pose}, when performing a plank, the pressure information provides information about the relative relationship between the hands and feet, as well as the physical interaction with the ground. This enables our method to estimate a reasonable pose even in the absence of visual signal about the legs. In contrast, other methods either fail to correctly estimate the leg posture (CLIFF, VIBE) or result in unrealistic floating of the legs (SMPLer-X). In the following step poses, we demonstrate that methods such as CLIFF and SMPLer-X, which rely solely on visual information, can be misled by 2D images. Although both achieve good 2D alignment, they produce unrealistic contact between the legs and the ground, as well as unreasonable shifts in the center of gravity.

\begin{figure}[tbp]
\centering
\includegraphics[width=1\linewidth]{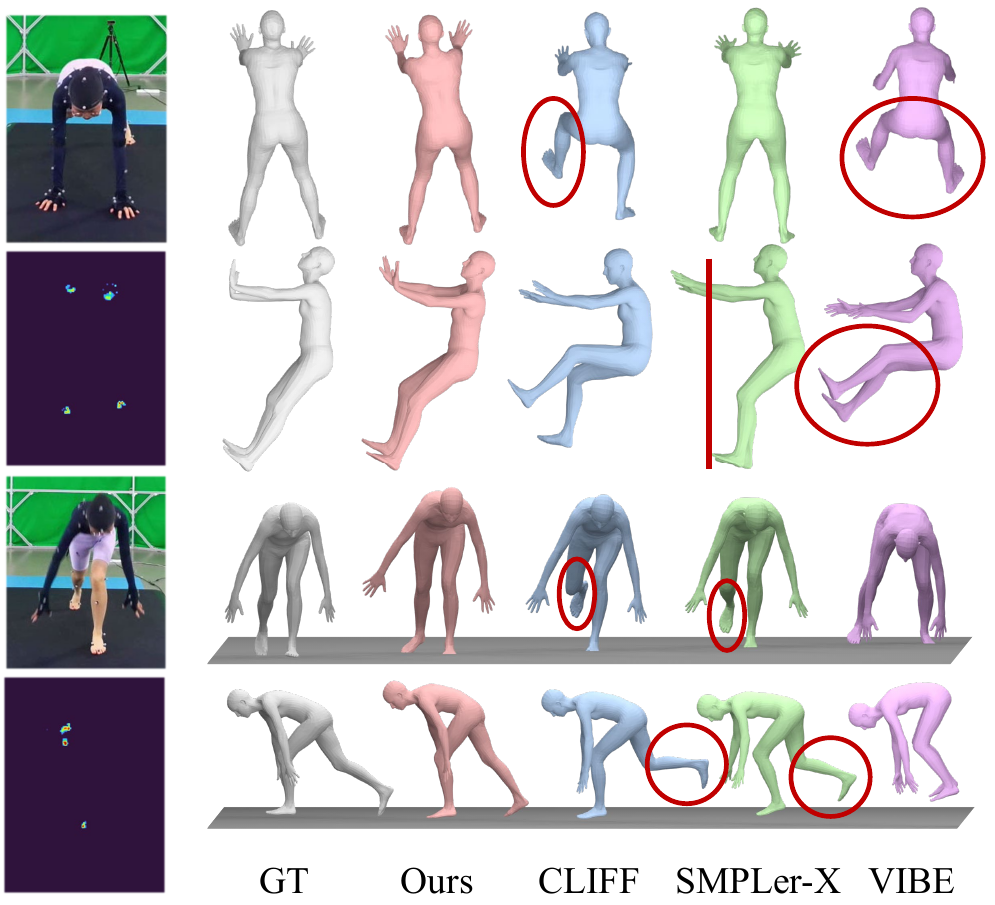}  
\caption{Qualitative comparison with methods for human pose estimation.}  
\label{fig:wpp}   
\vspace{-1em}
\end{figure}

\begin{figure}[tbp]
\centering
\includegraphics[width=1\linewidth]{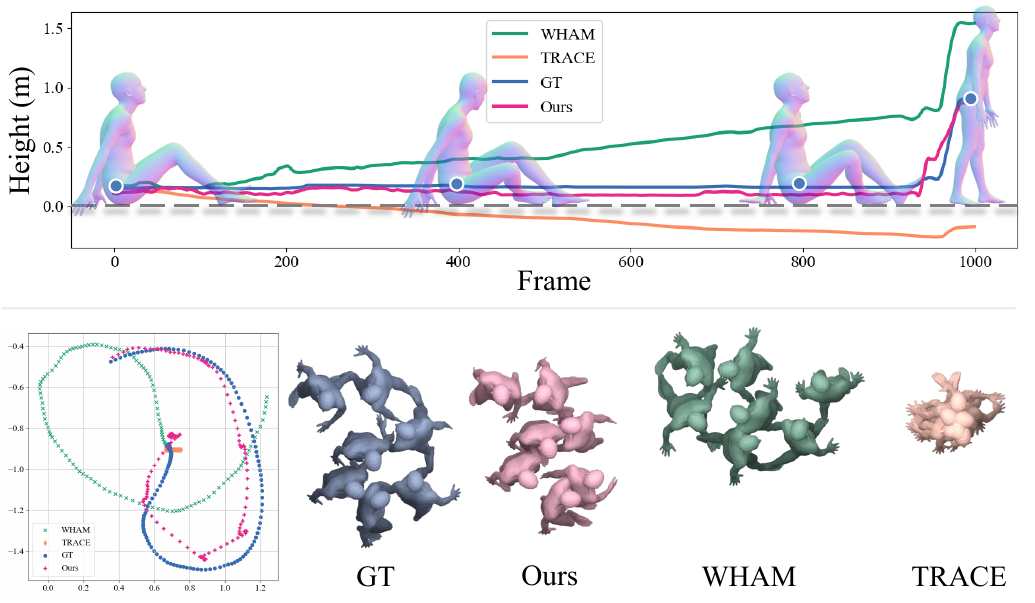}  
\caption{Qualitative comparison for global trajectory estimation.}  
\label{fig:visual_trans}  
\vspace{-0.5em}
\end{figure}

\noindent\textbf{Evaluation in global trajectory estimation.}
We compare FRAPPE with global trajectory estimation method WHAM~\cite{shin2024wham} and TRACE~\cite{sun2023trace} in the MotionPRO dataset. As shown in Tab.~\ref{tab:trans_evaluation}, our method outperforms all metrics.
For qualitative comparison, we evaluate the global trajectory in the vertical direction during a sitting and stand-up pose. As shown in Fig.~\ref{fig:visual_trans} , we plot the root joint height curves of different methods over time, after root joint alignment at the first frame. 
When the person is in a sitting position, we can ensure that the root joint maintains a trajectory consistent with the ground truth in terms of height, while both WHAM and TRACE exhibit upward or downward drift. Due to the physically plausible whole-body contact provided by pressure, we are able to achieve a reasonable relative positional relationship between the estimated human body and the ground. Due to the limitation of the dataset used by WHAM, which lacks motion types involving full-body contact with the ground, WHAM only considers the contact constraints of the feet and neglects the contact constraints of other body parts, such as the hips, hands, elbows, and knees. TRACE does not account for any relative positional relationships between the human body and the environment. Meanwhile, we evaluate the trajectory in the horizontal direction. Experiments show that our method still outperforms the other two methods in terms of global trajectory along the horizontal direction. 

\begin{table}[tbp]
  \centering
  \resizebox{1\linewidth}{!}{
  \begin{tabular}{@{}lccccc@{}}
    \toprule
    Methods & WMPJPE $\downarrow$ & WAMPJPE $\downarrow$ & RTE $\downarrow$ & Jitter  $\downarrow$ & WBCE $\downarrow$  \\
    \midrule
    TRACE~\cite{sun2023trace} & 141.2 & 92.5 & 1193 & 68.6   &10272.4 \\

    WHAM~\cite{shin2024wham} & 75.6 & {50.2}  & 1023 & 9.2  & 1217.6  \\
    \midrule
    Ours  & \cellcolor{First}{60.8} & \cellcolor{First}44.6 & \cellcolor{First}41.6 & \cellcolor{First}6.0  & \cellcolor{First}{110.2}  \\
    
    \bottomrule
  \end{tabular}
  }
  \vspace{-0.5em}
  \caption{Evaluation of global trajectory on MotionPRO. }
  \vspace{-0.5em}
  \label{tab:trans_evaluation}

\end{table}

\subsection{Ablation Study}

We conduct ablation studies on each of our crucial modules and loss functions. As shown in Tab.~\ref{tab:ablation}, our method performs best in global trajectory and global MPJPE, which is consistent with our motivation to drive virtual human or even humanoid robot in 3D scene. When FCAM is absent, MPJPE performs better than our method. This is because, without the fusion of visual and pressure signals, the network lacks access to physical dynamic information. As a result, the network tends to focus more on learning local pose rather than global pose and orientation. When 2D loss is omitted, metrics such as jitter and FS, which assess the physical plausibility of the motion, show better performance. This is because, in the absence of the 2D orthogonal projection constraint from the horizontal direction, the network becomes more focused on exploring the global physical information provided by the pressure signals. This shift leads to more realistic motions, as evidenced by smaller jitter and reduced foot sliding during contact. This also validates our insight that to achieve more plausible motions, a trade-off between 2D loss and 3D loss is necessary.

\begin{table}[tbp]
  \centering
  \resizebox{1\linewidth}{!}{
  \begin{tabular}{@{}lccccc@{}}
    \toprule
    Models & GTraj $\downarrow$ & GMPJPE $\downarrow$ & Jitter $\downarrow$  & MPJPE  $\downarrow$ & FS $\downarrow$  \\
    \midrule
    w/o LSAM & 93.1 & 92.9 & 6.3 & 44.4 &  3.6 \\

    w/o FCAM & 66.0 & 70.7 & 6.6 & \cellcolor{First}39.5    & 3.9 \\

    w/o contact loss & 68.7 & 75.9 & 8.5 & 42.6    & 5.7 \\
    % w/o contact and 2d & 72.7 & 82.3 & 8.8 & 43.5    & 6.0 \\
    w/o 2d loss & 82.1 & 95.5 & \cellcolor{First}5.5 & 52.2    & \cellcolor{First}3.3 \\

    \midrule
    Ours  & \cellcolor{First}{62.2} & \cellcolor{First}68.6  & 6.0& 40.5 & {3.6}  \\
    
    \bottomrule
  \end{tabular}
  }
  \vspace{-0.5em}
  \caption{Ablation study of FRAPPE. }
  \label{tab:ablation}
\vspace{-1em}
\end{table}

\section{Extended Application on Humanoid Robot}

In the field of embodied intelligence, generating reasonable and human-like robot movements is crucial, and using human motions to drive robots provides an efficient solution.
There have been several promising attempts in human-to-humanoid teleoperation and motion tracking systems~\cite{darvish2023teleoperation,fu2024humanplus, he2024learning}. However, robot motion tracking, particularly lower-body tracking, still remains inaccurate and unstable due to the lack of environmental interaction information and the inherent ambiguity in vision-based human pose estimation. 
Benefit from the accurately estimated pose and the contact information from the pressure data, we can further extend our proposed method to humanoid robot actuation. We first conduct a quantitative assessment in an ideal physical simulation environment, given the complexity of obtaining relevant state measurements for real-world humanoid robots. We then show the actuation of a humanoid robot using our method in a real environment.

\begin{figure}[tbp]
\centering
\includegraphics[width=1.0\linewidth]{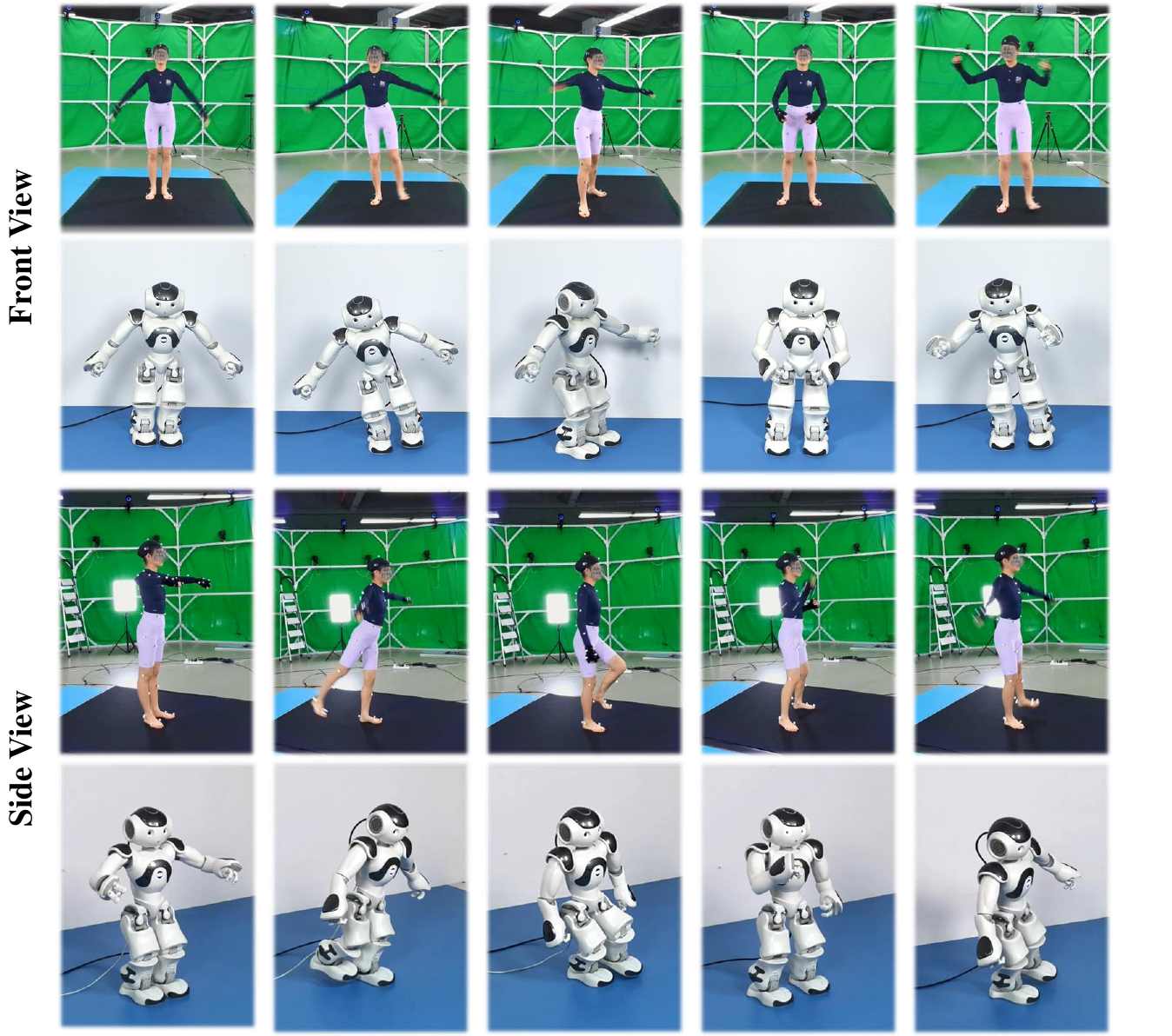}  
\caption{Motion actuation on a real robot}  
\label{fig:robot_imitation}   
\end{figure}

\begin{table}[tbp]
\begin{center}
\resizebox{0.49\textwidth}{!}{
    \begin{tabular}{ccccccc}
    \hline
    & \(\text{MPJPE-H}\downarrow\) & \(\text{MPJPE-R}\downarrow\) & \(\text{Fréchet}\downarrow\) & \(\text{Complete} \uparrow\) &\(E_\text{com}\downarrow\) &  \(E_\text{cop} \downarrow\)  \\
    \hline
    Optical  & 0 & 26.42 & 566.36 & 46.84 & 30.13 & 36.50 \\
    \hline
    CLIFF  & 83.60 & 19.10 & 574.52 & 65.50 & 33.32 & 35.84 \\
    CLIFF-P & 84.65 & 16.77 & 564.20 & 69.12 & 17.38 & 22.44 \\
    \hline

    Ours & \cellcolor{First}{65.11} & \cellcolor{First}{15.98} & 569.17 & 65.07 & \cellcolor{First}{15.32} & \cellcolor{First}{22.35} \\
    \hline
    \end{tabular}
}
\caption{Quantitative results of humanoid actuation.}
\label{tab:humanoid}
\end{center}
\vspace{-2.5em}
\end{table}

\noindent\textbf{Evaluation in physical simulation. }
We use the humanoid robot NAO and the physical simulation environment Webots to conduct the experiments. Specifically, we first estimate the human body's SMPL model from RGB images by FRAPPE and detect the bounding box of the feet from the pressure mat. Then, we implement a sensor fusion framework that integrates pressure with the RGB-derived pose~\cite{lu2024leveraging}, producing optimized foot articulation poses that better aligned with observed contact dynamics.
Following existing work~\cite{he2024learning,lu2024leveraging,zhang2022kinematic,tessler2024maskedmimic}, we use the metrics MPJPE-H, MPJPE-R (mm) to evaluate accuracy of human pose estimation and robot motion tracking, and Fréchet (mm) distance to evaluate motion similarity between human and robot. The percentage of time during which the robot successfully imitates motions without falling relative to the total duration of the motion sequence is to assess the completeness (\%). Additionally, we measure the mean global deviation \(E_\text{com}\) (mm) between the Center of Mass (CoM) projection and the ideal support region, as well as the deviation \(E_\text{cop}\) (mm) between the Center of Pressure (CoP) projection and the ideal support region, to evaluate the stability of humanoid motion. 

As shown in Tab.~\ref{tab:humanoid}, our method outperforms the baseline CLIFF and even surpasses the CLIFF pose refined by pressure (CLIFF-P) in terms of accuracy (MPJPE-H, MPJPE-R) and stability (\(E_\text{com}\),\(E_\text{cop}\)). Results demonstrate that the introducing pressure improves the accuracy and stability of the humanoid pose. Furthermore, compared to methods that use separated CLIFF and pressure as inputs without fusion, our FRAPPE method effectively fuses pressure and RGB to achieve a more accurate pose.
It is worth noting that due to the robot's limited structure and execution capabilities, higher accuracy in estimated human poses contrarily lead to a decrease in the similarity (Fréchet) between the robot's and human's movements.
Higher sensor accuracy may also lead to robot collapse during frequent support leg switching, resulting in a corresponding decrease in completeness. However, low-accuracy methods cannot accurately detect the switching conditions, thereby paradoxically maintaining higher completeness.

\noindent\textbf{Motion actuation in real environment. }
The real humanoid robot actuation is shown in Fig.~\ref{fig:robot_imitation}.
The humanoid can perform corresponding actions based on human performance.
Moreover, with the introduction of pressure information, the robot can more precisely detect changes in contact, enabling fine-grained lower-body motion imitation.

\vspace{-1em}

\section{Conclusion}

We construct the MotionPRO dataset, a large-scale multi-modal collection that integrates pressure, RGB, and optical sensors. We also propose FRAPPE, a novel baseline that combines pressure and RGB data to enhance pose and trajectory estimation. Through extensive experiments, we demonstrate that pressure signals not only improve the plausibility of lower-body pose estimation but also significantly enhance global trajectory prediction. Furthermore, we show that integrating pressure signals into humanoid robot actuation can stabilize and refine lower-body motion. Consequently, we explore the necessity of incorporating dynamic interaction mechanisms, such as pressure, into human motion capture systems. This work provides a rich resource for advancing motion capture research and opens up promising directions for future research in motion capture, augmented reality, and humanoid robotics.

\noindent\textbf{Acknowledgements:} This work is supported in part by the NSFC (No.62422110, No.82441013), Guoqiang Institute of Tsinghua University (No.2021GQG0001).

{
    \small
    \bibliographystyle{ieeenat_fullname}
    \bibliography{main}
}

\clearpage
\setcounter{page}{1}
\maketitlesupplementary

\section{Dataset Details:}
\noindent\textbf{Implementation Details.}

Each participant performs almost all the motion types. Each motion type is repeated two or three times. Each sequence represents a Sub-Motion Type in \cref{fig:motion_type} and lasts about 10 minutes. Following Human3.6M, we split the dataset into training and test sets at a 5:1 ratio based on participants, ensuring that there is no overlap between training and test sets for any \textless Participant, Motion Type \textgreater pair.

\noindent\textbf{Volunteers Details.}

\begin{figure}[htbp]
    \centering
    \includegraphics[width=0.6\linewidth]{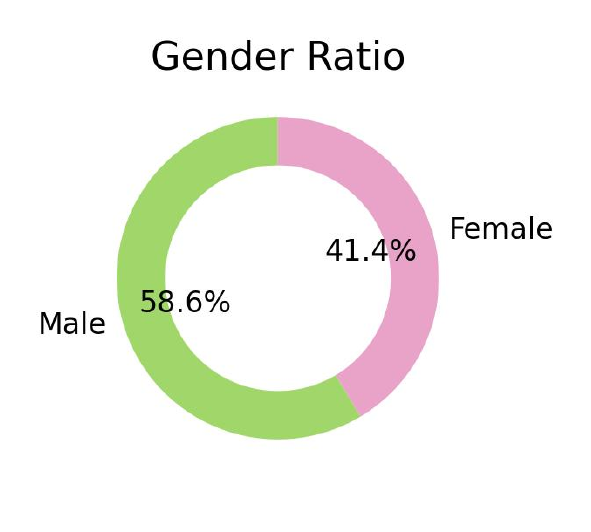}
    \caption{Gender Ratio of MotionPRO}
    \label{fig:gender}
\end{figure}

\textbf{Gender}: Our dataset consists of 70 individuals, comprising 29 females and 41 males, as shown in Fig.~\ref{fig:gender}.

\begin{figure}[htbp]
\centering
\includegraphics[width=1\linewidth]{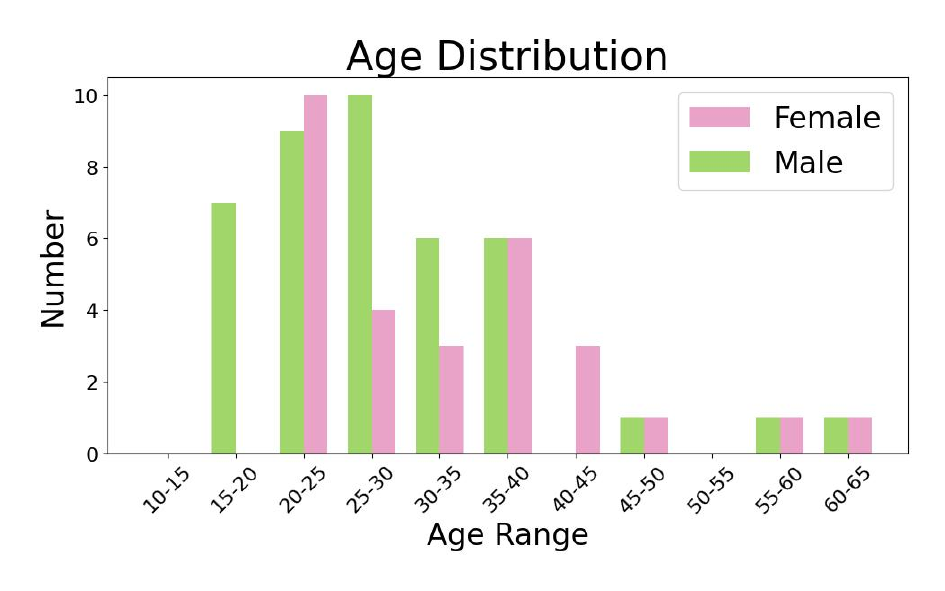}  
\caption{Age Distribution by Gender (5 years intervals)}  
\label{fig:ages}   
\end{figure}

\textbf{Age}: As shown in Fig.~\ref{fig:ages}, our dataset encompasses individuals across a broad age range, spanning from 15 to 61 years, with an average age of 31.4 for women and 26.6 for men.

\begin{figure}[htbp]
    \centering
    \includegraphics[width=1\linewidth]{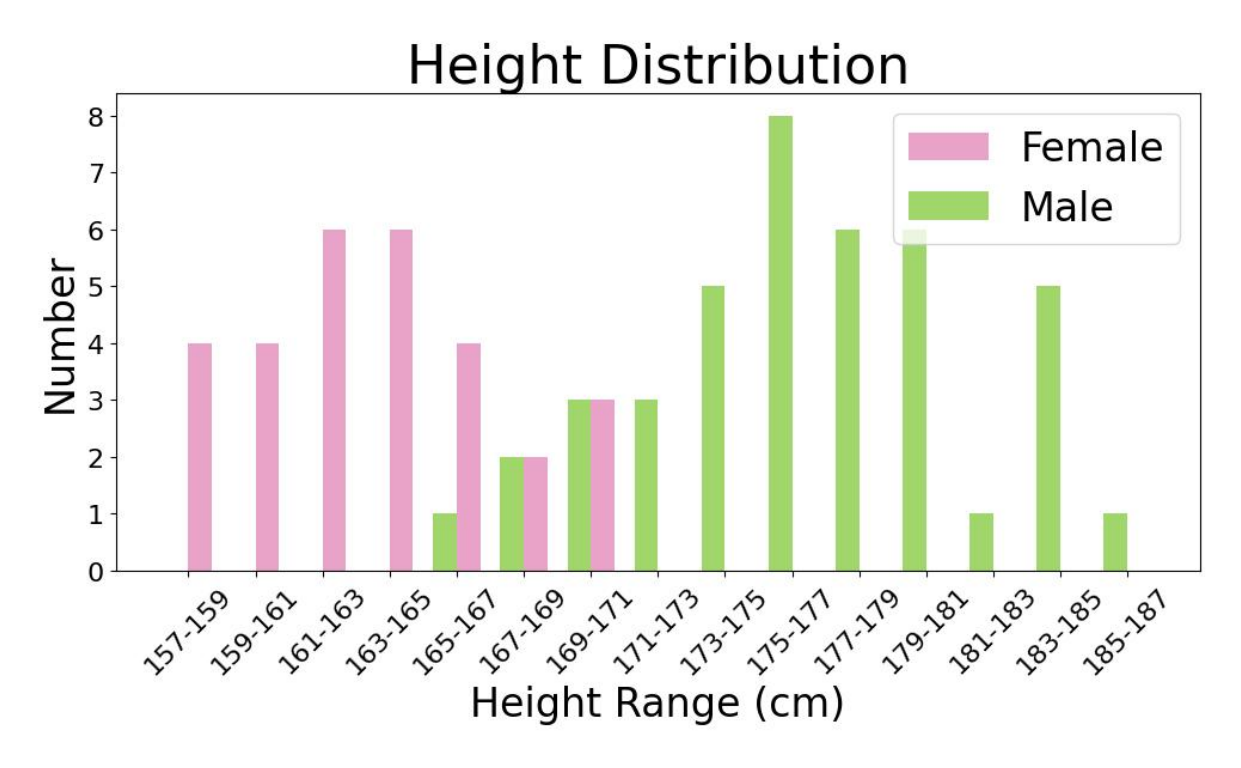}
    \caption{Height Distribution by Gender (5 cm intervals)}
    \label{fig:height}
\end{figure}

\textbf{Height}: As shown in Fig.~\ref{fig:height}, our dataset includes individuals of varying heights, spanning from 157 $cm$ to 185 $cm$, with an average height of 162.9 $cm$ for women and 176.2 $cm$ for men.

\begin{figure}[htbp]
    \centering
    \includegraphics[width=1\linewidth]{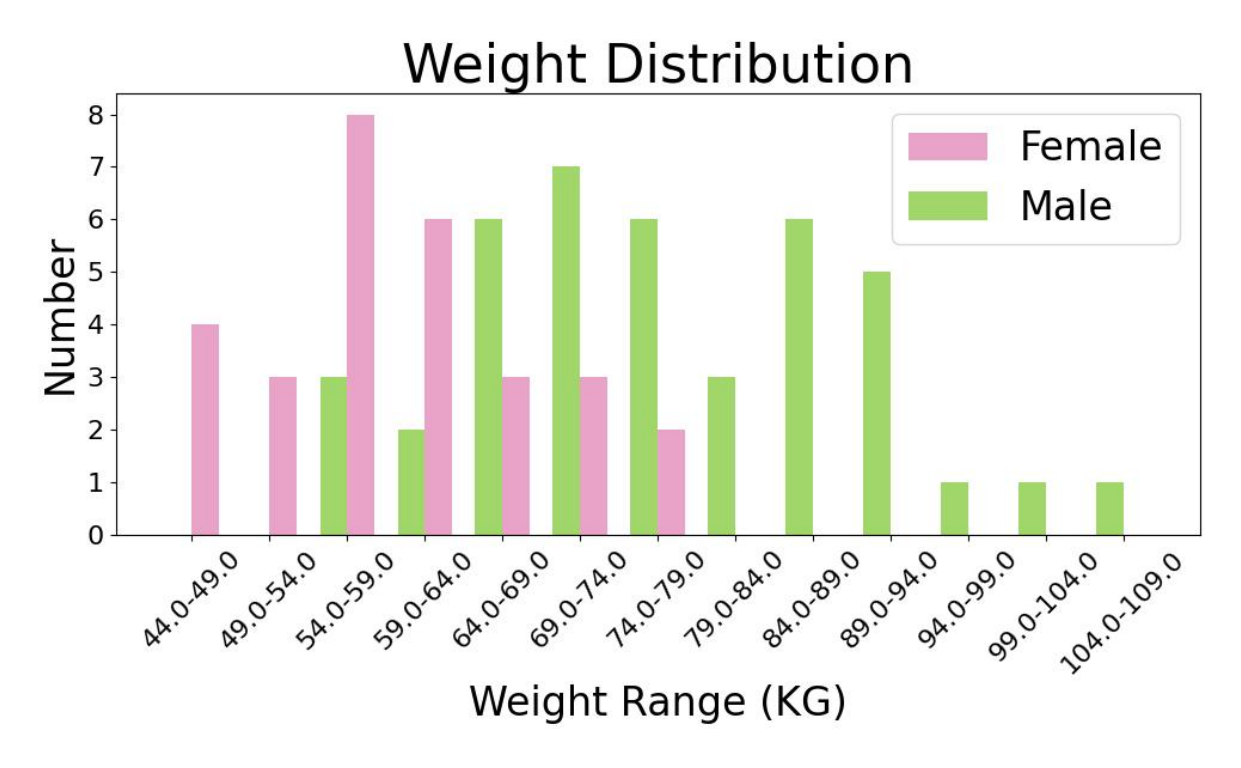}
    \caption{Weight Distribution by Gender (5 KG intervals)}
    \label{fig:weight range}
\end{figure}

\textbf{Weight}: As shown in Fig.~\ref{fig:weight range}, our dataset includes individuals with a range of weights, spanning from 44.1 $kg$ to 108 $kg$, with an average weight of 59.8 $kg$ for women and 78.0 $kg$ for men.

\noindent\textbf{Sensor Details.}

Our system utilizes a multi-sensor setup for data acquisition: 
\begin{itemize}
    \item 4 Azure Kinect cameras~\cite{azure2024kinect} to capture high-quality RGB videos. 
    \item 12 optical cameras (SWIFT 30)~\cite{luster2024fzmotion} to record raw marker data for precise motion tracking.  
    \item 1 pressure mat, specially designed for our system, to measure whole-body pressure during various motions.  
\end{itemize}

\noindent\textbf{Motion Types.}

\begin{figure}[htbp]
\centering
\includegraphics[width=1\linewidth]{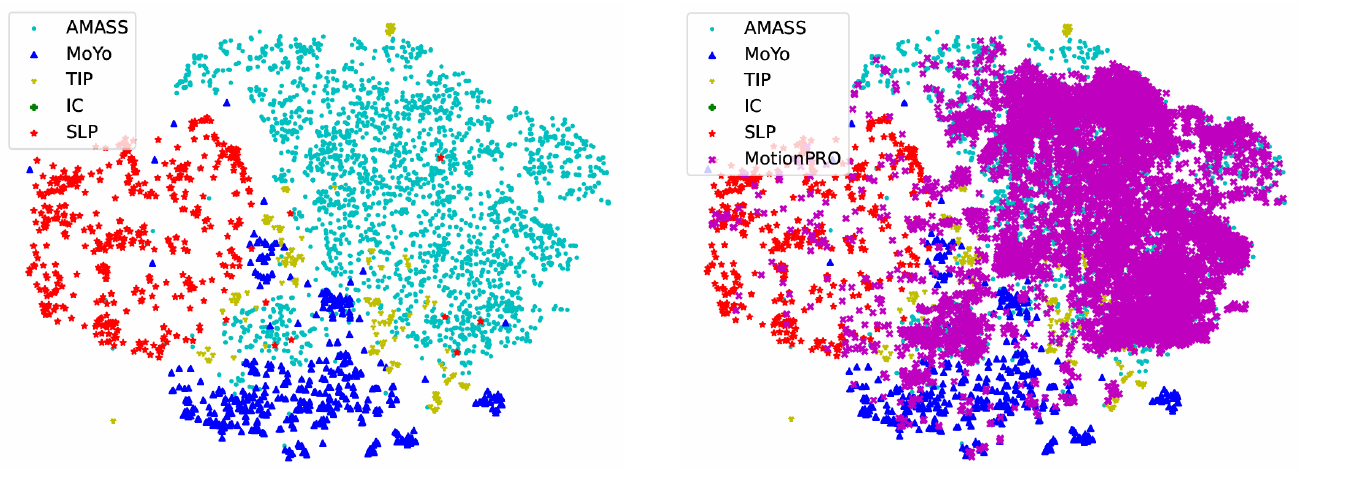}  
\caption{The distribution of poses in MotionPRO and existing MoCap datasets is visualized using T-SNE~\cite{van2008visualizing} dimensionality reduction. }  
\label{fig:tsne}   
\end{figure}

\begin{figure}[H]
\centering
\includegraphics[width=1\linewidth]{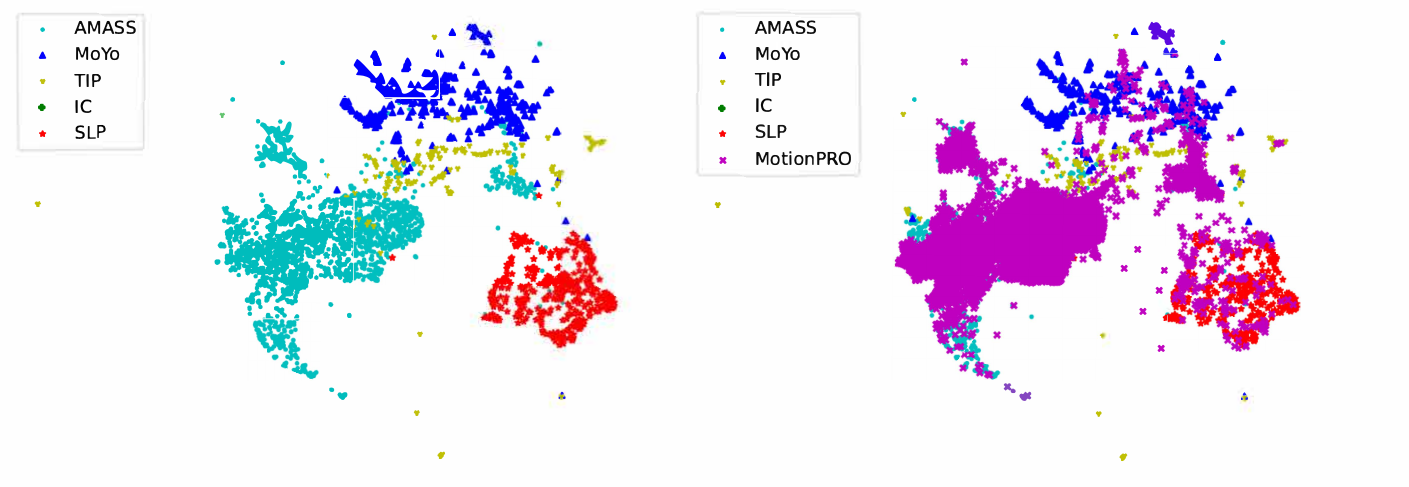}  
\caption{The distribution of poses in MotionPRO and existing MoCap datasets is visualized using UMAP~\cite{mcinnes2018umap} dimensionality reduction.}  
\label{fig:umap}   
\end{figure}

The T-SNE~\cite{van2008visualizing} and UMAP~\cite{mcinnes2018umap} plot in Fig.\ref{fig:tsne} and Fig.\ref{fig:umap} demonstrates that MotionPRO encompasses a wide range of motion types, nearly equivalent to the combined distribution of all currently available datasets (AMASS~\cite{mahmood2019amass}, MoYo~\cite{tripathi20233d}, TIP~\cite{wu2024seeing}, IC~\cite{luo2021intelligent}, SLP~\cite{liu2022simultaneously}). The figure on the left represents the T-SNE or UMAP distribution of the existing dataset, while the figure on the right illustrates the results of directly mapping MotionPRO based on the T-SNE or UMAP distribution observed on the left.

\noindent\textbf{Motion Categories.}

We define the six first-level categories as follows:

\textbf{Daily}: This category includes 172 common motions of daily life, such as basic postures, simple activities, and repetitive behaviors. These motions are characterized by natural, non-specialized patterns with high frequency, serving as a crucial baseline for developing human motions.

\textbf{Robot}: This category includes motions that simulate robotic or mechanized behaviors, characterized by mechanical patterns, fixed postures, high repetition, and predictability. Such data are essential for research on robotic motion simulation and human-robot interaction dynamics.

\textbf{Flexibility Exercise}: This category primarily includes motions involving large joint ranges of motion and the maintenance of slow, stable postures, such as leg stretches and splits.

\textbf{Aerobic Exercise}: This category comprises fitness activities defined by high-frequency, large-amplitude, full-body movements, typically associated with cardiovascular training.

\textbf{Traditional Chinese Exercise}: This category emphasizes movements characterized by fluidity, control, and balance, contrasting with high-intensity workouts and reflecting the characteristics of traditional Chinese fitness practices.

\noindent\textbf{Ethics.}

Volunteers in the MotionPRO dataset are well informed, and all participants have signed a Data Release Commitment Agreement, permitting the use of their data for research purposes.

\section{Baseline Details:}

\noindent\textbf{Intuition of pose estimation from pressure}

\begin{figure}[h]
  \centering
   \includegraphics[width=1.0\linewidth]{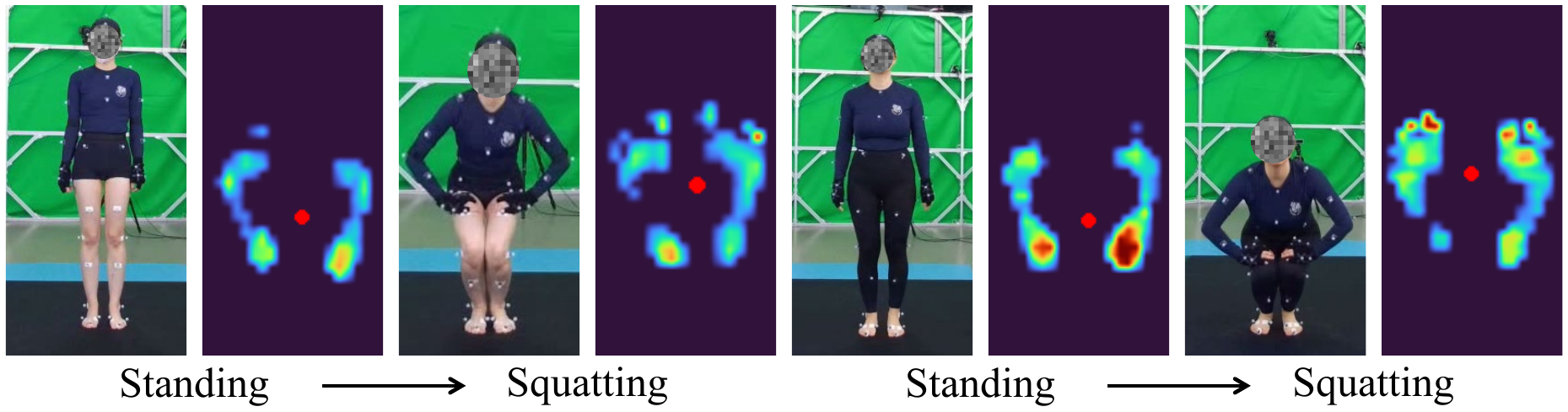}
   
   \caption{Comparison of pressure between standing and squatting. 
   % From left to right: SMPLer-X, TRACE, CLIFF, our method.
   }
   \label{fig:rebuttal}
\end{figure}

Through the spatial distribution and temporal changes of pressure, we verify that foot-to-floor pressure sensor readings can provide important discriminative prior information for pose estimation. Take standing and squatting as an example (shown in Fig.\ref{fig:rebuttal}), the CoP (Center of Pressure) is close to the heel and the toes exert almost no pressure on the ground when a person is standing. Conversely, when squatting, the CoP shifts closer to the forefoot and the toes generate pressure on the ground, helping to maintain balance. Additionally, the temporal relationship can provide more distinctive features. For example, when the posture transitions from standing to squatting, the body generates vertical acceleration, which leads to changes in both the total pressure value and the pressure distribution over time.

\noindent\textbf{Pressure network details.}

When standing, the effective pressure area is small, requiring more fine-grained feature extraction. To address this, we reduce the size of the first convolution kernel in the pressure encoder, enabling us to capture more features within the limited pressure area. LSAM comprises two layers of bidirectional GRU and one layer of Self-Attention, with each layer incorporating a residual connection. The specific configuration of the network structure is determined through testing on toy examples.

\noindent\textbf{Loss functions.}

The loss of pose parameters $\mathcal{L}_{pose}$ is the mean squared error between the predicted $\boldsymbol{\theta}$ and ground-truth pose parameters $\tilde{\boldsymbol{\theta}}$.

\begin{equation}
\mathcal{L}_{pose}=\|\boldsymbol{\theta}-\tilde{\boldsymbol{\theta}}\|_{2}^{2}
  \label{eq:Lpose}
\end{equation}

The 3D joint loss, \(\mathcal{L}_{3d}\), is the mean squared error between the predicted joints $\boldsymbol{J}(\boldsymbol{\theta}, \boldsymbol{T})$ and ground-truth whole-body joints $\tilde{\boldsymbol{J}
}(\boldsymbol{\theta}, \boldsymbol{T})$, after performing pelvis alignment.

\begin{equation}
\mathcal{L}_{3d}=\|\boldsymbol{J}(\boldsymbol{\theta}, \boldsymbol{T}) - \tilde{\boldsymbol{J}
}(\boldsymbol{\theta}, \boldsymbol{T})\|_{2}^{2}
  \label{eq:L3d}
\end{equation}

Global translation loss $\mathcal{L}_{trans}$ is the mean squared error between predicted translation $\boldsymbol{T}$ and ground truth translation $\tilde{\boldsymbol{T}}$.

\begin{equation}
\mathcal{L}_{trans}=\|\boldsymbol{T} - \tilde{\boldsymbol{T}
}\|_{2}^{2}
  \label{eq:Ltrans}
\end{equation}

The ground contact loss, \(\mathcal{L}_{contact}\), is the mean squared error between the predicted global whole-body 
in-contact joints $\boldsymbol{J}_{C}(\boldsymbol{\theta}, \boldsymbol{T})$ and the ground-truth global whole-body in-contact joints $\tilde{\boldsymbol{J}}_{C}
(\boldsymbol{\theta}, \boldsymbol{T})$.

\begin{equation}
\mathcal{L}_{contact}=\| \boldsymbol{J}_{C}(\boldsymbol{\theta}, \boldsymbol{T}) - \tilde{\boldsymbol{J}}_{C}
(\boldsymbol{\theta}, \boldsymbol{T})\|_{2}^{2}
  \label{eq:Lcontact}
\end{equation}

$\mathcal{L}_{2d}$ is the mean squared error of orthographic projection $\mathcal{O}(\cdot)$ in the camera direction between the predicted joints and ground truth joints. 

\begin{equation}
\mathcal{L}_{2d}=\| \mathcal{O}(\boldsymbol{J}(\boldsymbol{\theta}, \boldsymbol{T}))-\mathcal{O}(\tilde{\boldsymbol{J}}(\boldsymbol{\theta}, \boldsymbol{T}))\|_{2}^{2}, 
  \label{eq:L2d}
\end{equation}

\noindent\textbf{Implement Details.}

When driving virtual humans or robots in a 3D environment, their shapes typically remain constant over time. These shapes are often specifically designed and can differ significantly from those of human motion providers. Therefore, human body shape estimation is not our focus. In both \textbf{Pose and Trajectory Estimation using Only Pressure} experiment and \textbf{Pose and Trajectory Estimation by Fusing Pressure and RGB} experiment, we do not utilize FRAPPE to estimate body shape. Instead, we pre-calculate a more reasonable and representative shape based on the actual human body dimensions and maintain it fixed throughout training and evaluation. Similarly, the shapes for other comparison methods are also set to a consistent shape to ensure fairness in evaluation. FRAPPE outputs the SMPL pose and translation parameters $\boldsymbol{\theta}, \boldsymbol{T}$.

FRAPPE takes \( 20 \) frames of consecutive RGB and pressure images as input. The RGB images used in our method are captured from a frontal view monocular camera, providing a direct perspective for motion analysis. Notably, in the image branch, the encoder parameters are kept frozen during training. This ensures that the model focuses on learning the fusion of pressure and RGB features rather than relearning image-specific features. At the same time, we also ensure fairness in comparison with other methods on the MotionPRO dataset, that is, our RGB image encoder, like other methods, is not trained on the MotionPRO dataset. We use AdamW optimizer with an initial learning rate of $5e^{-5}$ on 4 RTX 4090D GPUs. 

\section{Robot Actuation Details:}
\begin{figure}[t]
\centering
\includegraphics[width=1\linewidth]{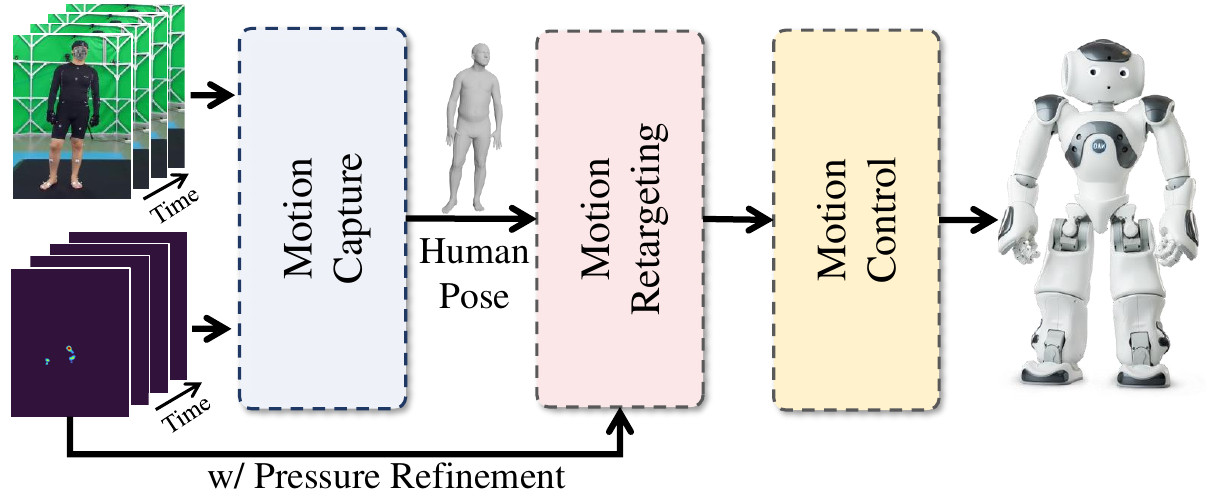}  
\caption{Framework of the robot demonstration system.}  
\label{fig:robot_system}   
\end{figure}

% \noindent\textbf{Implementation.}
We use the estimated human pose to actuate the robot. Our robot demonstration system is shown in Fig.~\ref{fig:robot_system}. Specifically, we first extract human skeletal joint points from the SMPL model, which is estimated in motion capture module. The human joint points are then retargeted to corresponding target joint points that the robot can execute, involving coordinate transformation, scaling, Center of Mass (CoM) tracking, and other related processes. Finally, in the robot motion control module, we provide the retargeted pose to the robot controller for inverse kinematics optimization and whole-body control. For further details, refer to ~\cite{koenemann2014real,otani2017adaptive,penco2018robust,lu2024leveraging}.

Through the analysis of our framework, we argue that the performance of the robot's action depends not only on the motion capture module but also on the other modules. Therefore, we investigate further optimization of the motion retargeting modules through the use of pressure data. Specifically, as the CoM distribution of the estimated human model does not perfectly align with the real pressure data, we refine the joint points using the pressure data, following \cite{lu2024leveraging}, to ensure that the body CoM offset aligns with the pressure offset. Moreover, pressure data provides highly accurate information on human body contact, which can be used as a reference for controlling the robot's support mode.
We apply this approach to CLIFF and FRAPPE and corresponding results are shown in the main text.

We now clarify why CLIFF method performs better than ours in completeness, as discussed in the main text. For challenging actions that the robot cannot perform in the dataset, such as jumping, lying down, and the plank pose, etc, our method leads to the robot falling when imitating due to the higher accuracy of our estimated poses. In contrast, CLIFF’s less accurate poses allow the robot to remain standing and continue demonstrating the next action.
In addition, it should be mentioned that the MPJPE-H metric primarily measures the difference between the estimated human pose and the ground truth. As we use the human pose captured by the optical system as the ground truth, resulting in a value of 0 for the optical MPJPE-H in Tab. 6 of the main text.

\section{Future Work}

Our dataset offers valuable opportunities for future research, particularly to examine the relationship between contact duration within the Base of Support (BoS), the distance between the Center of Mass (CoM) and the Center of Pressure (CoP), and demographic factors such as age, weight, and height. In addition, it supports applications in health monitoring and sports training. A key next step is to infer pressure information from visual input, which would expand its applicability by reducing the reliance on specialized sensors. Our dataset provides essential support for the advancement of these research directions.

% {
%     \small
%     \bibliographystyle{ieeenat_fullname}
%     \bibliography{main}
% }

\end{document}